\documentclass[letterpaper, 10 pt, journal, twoside]{IEEEtran}
\usepackage{cite}

\ifCLASSINFOpdf
   \usepackage[pdftex]{graphicx}
\else
   \usepackage[dvips]{graphicx}
\fi
\usepackage{amsmath}
\usepackage{algorithmic}

\usepackage{array}

\usepackage{xcolor}

\ifCLASSOPTIONcompsoc
 \usepackage[caption=false,font=normalsize,labelfont=sf,textfont=sf]{subfig}
\else
 \usepackage[caption=false,font=footnotesize]{subfig}
\fi

\ifCLASSOPTIONcaptionsoff
 \usepackage[nomarkers]{endfloat}
\let\MYoriglatexcaption\caption
\renewcommand{\caption}[2][\relax]{\MYoriglatexcaption[#2]{#2}}
\fi
\usepackage{url}
\usepackage[hidelinks]{hyperref}

\usepackage{glossaries}
\setacronymstyle{long-short}

\usepackage{eso-pic}

\AddToShipoutPictureBG*{%
   \AtPageUpperLeft{\put(3,-10){\scriptsize This article has been accepted for publication in a future issue of this journal, but has not been fully edited. Content may change prior to final publication. Citation information: DOI 10.1109/LRA.2021.3090016,}}
   \AtPageUpperLeft{\put(250,-17){\scriptsize IEEE Robotics and Automation Letters}}
   \AtPageLowerLeft{\put(3,10){\scriptsize 2377-3766 \copyright~2021 IEEE. Personal use is permitted, but republication/redistribution requires IEEE permission. See \url{http://www.ieee.org/publications_standards/publications/rights/index.html} for more information.}}
}

\begin{document}
%
\title{Position-based Dynamics Simulator of\\Brain Deformations for Path Planning and Intra-Operative Control in Keyhole Neurosurgery}
%
%
%

\author{Alice~Segato$^{\dag, 1}$, Chiara~Di~Vece$^{\dag, 1}$, Sara~Zucchelli$^{1}$, Marco~Di~Marzo$^{1}$, Thomas~Wendler$^{2}$, Mohammad~Farid~Azampour$^{2}$, Stefano~Galvan$^{3}$, Riccardo~Secoli$^{3}$ and Elena~De~Momi$^{1}$%
\thanks{Manuscript received: February, 24, 2021; Revised April, 22, 2021; Accepted June, 8, 2021.}
\thanks{This paper was recommended for publication by Editor Pietro Valdastri upon evaluation of the Associate Editor and Reviewers' comments.
This work was supported by the European Union's EU Research and Innovation programme Horizon 2020 under Grant agreement 688279.} 
\thanks{$^{\dag}$Alice Segato and Chiara Di Vece are co-first authors.}
\thanks{$^{1} $Alice Segato, Chiara Di Vece, Sara Zucchelli, Marco Di Marzo and Elena De Momi are with the NearLab, Department of Electronics, Information and Bioengineering, Politecnico di Milano, 20133 Milan, Italy
        {\tt\footnotesize \{alice.segato,elena.demomi\}@polimi.it~\{chiara.divece, sara.zucchelli,marco.dimarzo\}@mail.polimi.it}}%
\thanks{$^{2} $Thomas Wendler and Mohammad Farid Azampour are with Computer-Aided Medical Procedures and Augmented Reality, Technical University of Munich, Munich, Germany
        {\tt\footnotesize \{wendler,mf.azampour\}@tum.de}}%
\thanks{$^{3} $Stefano Galvan and Riccardo Secoli are with the Mechatronics In Medicine Laboratory, Mechanical Engineering Department, Imperial College London, London SW7 2BU, U.K.
        {\tt\footnotesize \{s.galvan, r.secoli\}@imperial.ac.uk}}%
\thanks{Digital Object Identifier (DOI): see top of this page.}
}
%
%

\markboth{IEEE Robotics and Automation Letters. Preprint Version. Accepted June, 2021}
{Segato, Di Vece \MakeLowercase{\textit{et al.}}: Position-based dynamics simulation of brain deformations in keyhole neurosurgery} 

%



\newacronym{pbd}{PBD}{Position-based Dynamics}
\newacronym{fem}{FEM}{Finite Element Method}
\newacronym{kn}{KN}{keyhole neurosurgery}
\newacronym{ai}{AI}{Artificial Intelligence}
\newacronym{rl}{RL}{Reinforcement Learning}
\newacronym{drl}{DRL}{Deep Reinforcement Learning}
\newacronym{mr}{MR}{Magnetic Resonance}
\newacronym{ct}{CT}{Computerized Tomography}
\newacronym{od}{OD}{outer diameter}
\newacronym{pbn}{PBN}{Programmable Bevel-Tip Needle}
\newacronym{com}{CoM}{center of mass}
\newacronym{csf}{CSF}{cerebrospinal fluid}
\newacronym{mch}{MCH}{composite hydrogel}
\newacronym{ffe}{FFE}{fast-field-echo}
\newacronym{tof}{TOF}{time‐of‐flight}
\newacronym{ai}{AI}{Artificial Intelligence}
\newacronym{mre}{MRE}{Magnetic Resonance Elastography}

\maketitle

\begin{abstract}
Many tasks in robot-assisted surgery require planning and controlling manipulators' motions that interact with highly deformable objects. This study proposes a realistic, time-bounded simulator based on \gls{pbd} simulation that mocks brain deformations due to catheter insertion for pre-operative path planning and intra-operative guidance in keyhole surgical procedures. It maximizes the probability of success by accounting for uncertainty in deformation models, noisy sensing, and unpredictable actuation. The \gls{pbd} deformation parameters were initialized on a parallelepiped-shaped simulated phantom to obtain a reasonable starting guess for the brain white matter. They were calibrated by comparing the obtained displacements with deformation data for catheter insertion in a composite hydrogel phantom. Knowing the gray matter brain structures' different behaviors, the parameters were fine-tuned to obtain a generalized human brain model. The brain structures' average displacement was compared with values in the literature. The simulator's numerical model uses a novel approach with respect to the literature, and it has proved to be a close match with real brain deformations through validation using recorded deformation data of in-vivo animal trials with a mean mismatch of 4.73$\pm$2.15\%. The stability, accuracy, and real-time performance make this model suitable for creating a dynamic environment for KN path planning, pre-operative path planning, and intra-operative guidance.
\end{abstract}


%
\IEEEpeerreviewmaketitle

\section{Introduction}
\label{sec:intro}
\glsresetall
%
%
%
%
\IEEEPARstart{K}{eyhole} neurosurgery (KN) is a minimally invasive procedure performed to reach targets located deep inside the brain through a tiny hole in the skull, called ``burr hole'' or ``keyhole''~\cite{keyhole}. In KN, catheters are inserted into the brain for biopsy and therapies, as drug delivery or electrical stimulation following predefined paths to avoid damage to delicate structures. 

Path planning plays an essential role in various fields of application and research. In its most general form, the task is to define a path for some moving entity between a start and a goal in an environment, {\em e.g.}, the trajectory of a catheter between keyhole and target location in an \gls{mr} image. Nonetheless, the difference between the pre-operative planning in a static condition where no deformation is considered and the intra-operative dynamic condition where the anatomy deforms can significantly impact the planned trajectory's accuracy and relevance, since both the target organs and the obstacles such as vessels can move~\cite{essert1}. Due to the difficulty,
tasks involving uncertainty and highly deformable objects are still routinely completed manually rather than automatically or semi-autonomously using robot assistance. Automating these tasks could increase productivity and improve outcomes by decreasing the time and costs associated with manual operation while simultaneously increasing accuracy and precision.

When considering the brain, there are three primary sources of deformation: the pulsation of brain vessels, the brain shifts~\cite{brainshift1,brainshift2,brainshift3,brainshift4} (which could be caused by (i) \gls{csf} loss through the keyhole, resulting in surgical accuracy deterioration, (ii) pharmaceuticals administered during the intervention and (iii) tissue resection~\cite{def}), and the interaction between the catheter and the brain tissue. However, the pulsation of the brain vessels is not relevant in minimally invasive surgery~\cite{pulsation2}. 

Synthetic phantoms and virtual models of the human brain have been proposed as a tool to measure/compute these deformations and compensate for them. Such models are essential to reproduce geometries and structures and provide material formulations that could accurately replicate such a complex organ's mechanical behavior. Many works in the literature have studied brain deformations during neurosurgery~\cite{def2,def3} and brain biomechanical characterization~\cite{imperial1,imperial2,imperial3,imperial4}. Most of the physics-based brain deformation models are based on \gls{fem} but differ from each other in the choice of the constitutive equation of the model that defines the material chosen for the soft tissue simulation. In particular,~\cite{imperial1} presents a comprehensive comparative study to help researchers select suitable materials for soft tissue simulation and their interactions with surgical instruments. The authors matched the stiffness of gelatin and \gls{mch} phantoms to the one of a porcine brain and analyzed needle insertions in terms of insertion forces, displacements, and deformations. In \cite{imperial2}, the authors perform a characterization of brain tissues under various test conditions (including humidity, temperature, strain rate); they present a rigorous experimental investigation of the human brain's mechanical properties through ex-vivo tests, covering both gray and white matter. Brain material characterization is instrumental in providing inputs to a mathematical model describing brain deformation during a surgical procedure. The work of Forte et al. \cite{imperial3} focuses on developing an accurate numerical model for predicting brain displacement during procedures and employs a tissue mimic of \gls{mch} to capture human tissue complexity. This mimic is designed to reproduce the dynamic mechanical behavior in a range of deformation rates suitable for surgical procedures. The investigation supports the proposal of a hybrid formulation of porous-hyper-viscoelastic material for brain displacement simulation. Further work of Leibinger et al.~\cite{imperial4} measures internal displacements and strains in three dimensions within a soft tissue phantom at the needle interface, providing a biologically inspired solution causing significantly fewer damages to surrounding tissues. 

As an alternative for the phantoms and models mentioned above, neurosurgical simulators based on \gls{fem} brain tissue deformation model have been developed. Some of these are based on the optimization of the implicit Euler method, such as the one presented in~\cite{hou2019new}. Others like the NeuroTouch~\cite{delorme2012neurotouch}, a commercial neurosurgical simulator, use the explicit time integration scheme, require small time steps to keep the computation stable. Implicit integration schemes~\cite{allard2012implicit}, on the other hand, have the advantage of unconditional stability but are much more computationally expensive~\cite{benderPBD}. 

Recently, \gls{pbd}~\cite{muller} has gained considerable popularity due to its simplicity, high efficiency, unconditional stability, and real-time performance~\cite{pbd}. \gls{pbd} approach eliminates the overshooting problem of equilibrium configurations achievement in force-based methods and simplifies the implementation process. Although \gls{pbd} is not as accurate as force-based methods, its efficiency and controllability outperform those simulating medical procedures by far while providing visually plausible results~\cite{camara}. 

To the authors' knowledge, this paper presents the first simulator of realistic brain tissue displacement using \gls{pbd} modeling designed for catheter insertion during KN. We further validate it using recorded brain deformation on an in-vivo animal model. The simulator developed in Unity with NVIDIA FleX~\cite{flex} backend provides a suitable system for \gls{ai} framework to train autonomous control and path planners. Moreover, we make our simulator publicly available on \href{https://gitlab.com/polimi-pathplanning/pbd-neurosurgical-simulator}{\underline{GitLab}} to encourage the adoption of the method. 
%
%
%
%
\section{Materials and Methods} 
\label{sec:methods} 
\subsection{Position-based dynamics approach}
\gls{pbd} is a simulation approach that computes the time evolution of a dynamic system by directly updating positions, as first described by M\"uller et al. in~\cite{mullerPBD}. Simulated objects are discretized as clusters of particles described by their positions $\textbf{p}_i$ and velocities $\textbf{v}_i$, subject to a set of $J$ positional constraints $C_j(\textbf{p}_i, ..., \textbf{p}_n)$~$\succ$~0 (symbol $\succ$ denotes either = or~$\ge$). In the \gls{pbd} approach, deformation computation becomes a constrained optimization problem. Soft bodies simulation behavior and performance are not only influenced by the relative position, dimension, and the number of particles in space but also by the constraints acting among particles. For example, large deformations of soft bodies are usually achieved by defining positional constraints among adjacent particles' rigid clusters. This kind of constraint is called \textit{region-based shape matching}, as described in~\cite{benderPBD}. A realistic elastic behavior is obtained by appropriately selecting cluster parameters; hence, we focused on the initialization of the \gls{pbd} parameters defining clusters for all the structures present in the scene: \textit{cluster spacing} ({\em i.e.,} the distance between adjacent clusters), \textit{cluster radius} ({\em i.e.,} the radius of each cluster region) and \textit{cluster stiffness} ({\em i.e.,} the extent to which adjacent clusters are constrained to each other). Other parameters of the \gls{pbd} model can impact soft body behavior and require tuning according to~\cite{muller}; therefore, to simplify the model, parameter values were kept fixed as in~\cite{camara}.
The NVIDIA FleX is a particle-based simulation library for real-time, realistic visual effects. The objects in the framework are modeled as clusters of particles connected by various constraints, broadly based on \gls{pbd}. Since NVIDIA FleX does not directly support the Unity Game Engine, we used uFleX, a Unity asset integrated low-level Flex native library.
\subsection{Case study: keyhole neurosurgery}
\begin{figure}[t]
    \centering
    \includegraphics[width=\columnwidth]{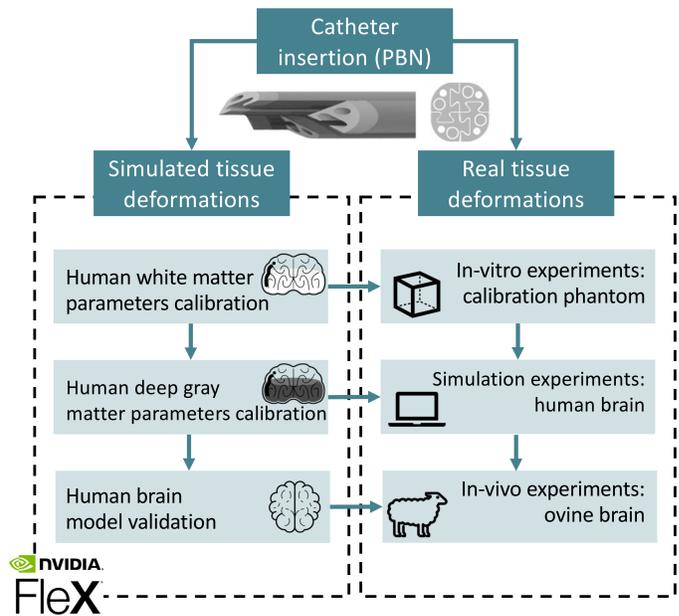}
    \caption{Workflow of the experimental study}
    \label{fig:figure1}
\end{figure}
Fig.~\ref{fig:figure1} summarizes the workflow of our experimental study made of two calibration phases, one for the white matter performed with an in-vitro experiment on a phantom and one for deep gray matter performed with an in-silico experiment on a human brain dataset. The last phase is the validation conducted through in-vivo experiments on ovine brains. 
The different steps are detailed below:
\subsubsection{Catheter}
\begin{figure}
    \centering
    \includegraphics[width=\columnwidth]{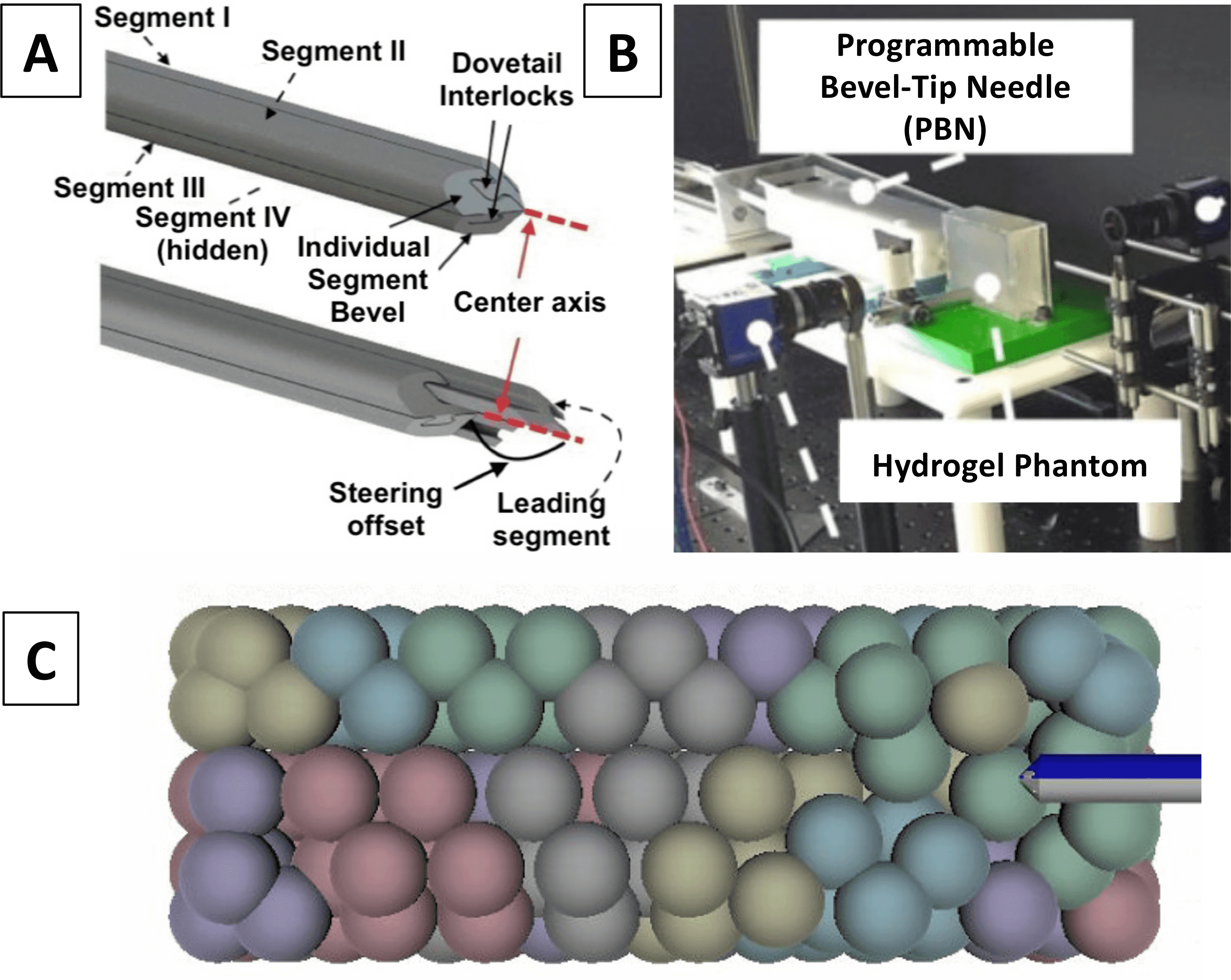} 
    \caption{\textbf{(A)} Illustration of the \gls{pbn} catheter showing the four segments interlocked by means of Dovetail joints. The four segments can slide on each other along the center axis, creating an offset that allows the steering of the catheter (courtesy of Leibinger, 2016). \textbf{(B)} Experimental setup with the hydrogel phantom and the needle presented in~\cite{imperial4}. \textbf{(C)} Basic deformable parallelepiped-shaped phantom used for the initial setting of \gls{pbd} simulation parameters for the brain white matter and the 3D model of the reconstructed catheter with an outer diameter of $2.5~mm$.}
    \label{fig:figure2}
\end{figure}

This project was developed in the context of the EU's RIA Horizon project codename EDEN2020, aiming
to advance the current state of the art in neurosurgical technology. The catheter model used in this work is a Programmable Bevel-Tip Needle (\gls{pbn})~\cite{eden}. The catheter has an outer diameter of 2.5~$mm$, as shown in Fig.~\ref{fig:figure2}A.
\subsubsection{Data acquisition and processing} 
\begin{figure}
    \centering
    \includegraphics[width=\columnwidth]{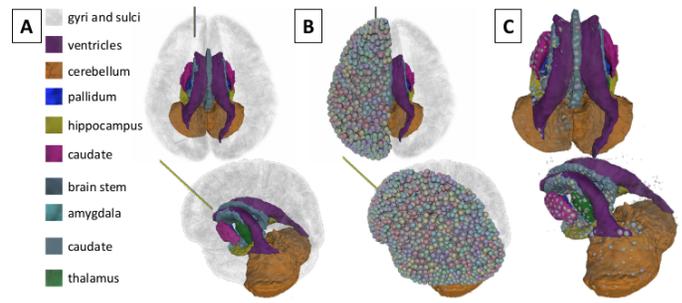} 
    \caption{\textbf{(A)} Representation of the brain structures segmented on the normalized 3D T1-weighted images and the 3D high‐resolution \gls{tof}-MRA. \textbf{(B)} Particle model of brain white matter. \textbf{(C)} Particle model of deep gray matter structures.}
    \label{fig:figure3}
\end{figure}

Various models were created to carry out our simulations:
\paragraph{Calibration brain phantom} 
We created a parallelepiped-shaped phantom (50$\times$17.5$\times$ 17.5~$mm$) using Blender~\cite{blender} to initialize the \gls{pbd} simulation parameters for the brain white matter. Its dimensions have been chosen to match the experimental setup of~\cite{imperial4}, shown in Fig.~\ref{fig:figure2}B. Fig.~\ref{fig:figure2}C shows the obtained simulated phantom and catheter. 
\paragraph{Human brain}
High-resolution \gls{mr} images were acquired on one healthy subject (male, aged 38 yo). The research ethical committee of Vita-Salute San Raffaele University and IRCCS San Raffaele Scientific Institute, Milan, Italy, approved the study (ethical approval n. 80/INT/2016), and the subject provided signed informed consent before \gls{mr} acquisition. The \gls{mr} protocol included:
\begin{itemize}
    \item a 3D T1-weighted sagittal \gls{ffe} with selective water excitation (Proset technique)~\cite{proset} acquired with the following parameters: repetition time/echo time [TR/TE] 12/5.9 $ms$; flip angle, 8$^{\circ}$; acquisition matrix, 320$\times$299; voxel size, 0.8$\times$0.8$\times$0.8 $mm$; thickness, 0.8/0 $mm$ gap; SENSitivity-Encoding [SENSE] reduction factor, R=2; 236 slices; acquisition time, 5~$min$ 19~$s$;  
    \item a 3D high‐resolution \gls{tof} \gls{mr} angiography (MRA) acquisition to visualize flow within the arterial vessels, acquired with the following parameters: TR/TE 23/3.45 $ms$; flip angle, 18$^{\circ}$; acquisition matrix, 500$\times$399; acquired voxel size, 0.4$\times$0.5$\times$0.9 $mm$; reconstructed voxel size, 0.3$\times$0.3$\times$0.45 $mm$; thickness, 0.45/-0.45 $mm$ gap; SENSE factor, R~=~2; 210 slices; acquisition time, 8~$min$ 33~$s$.
\end{itemize}
On the normalized 3D T1-weighted images, we segmented the amygdala, brain stem, caudate, cerebellum, cerebral cortex (gyri and sulci), hippocampus, pallidum, putamen, thalamus, and ventricles employing FreeSurfer Software~\cite{freesurfer,freesurfer2}. Additionally, we segmented the arterial blood vessels from the 3D high‐resolution \gls{tof}-MRA using 3D Slicer© platform~\cite{slicer,slicer2} to obtain the final set of brain structures ($BS = bs_1, ..., bs_m$, with $m = \#BS$). Fig.~\ref{fig:figure3}A shows the final result.
\paragraph{Ovine brain}
\begin{figure*}
    \centering
    \includegraphics[width=\textwidth]{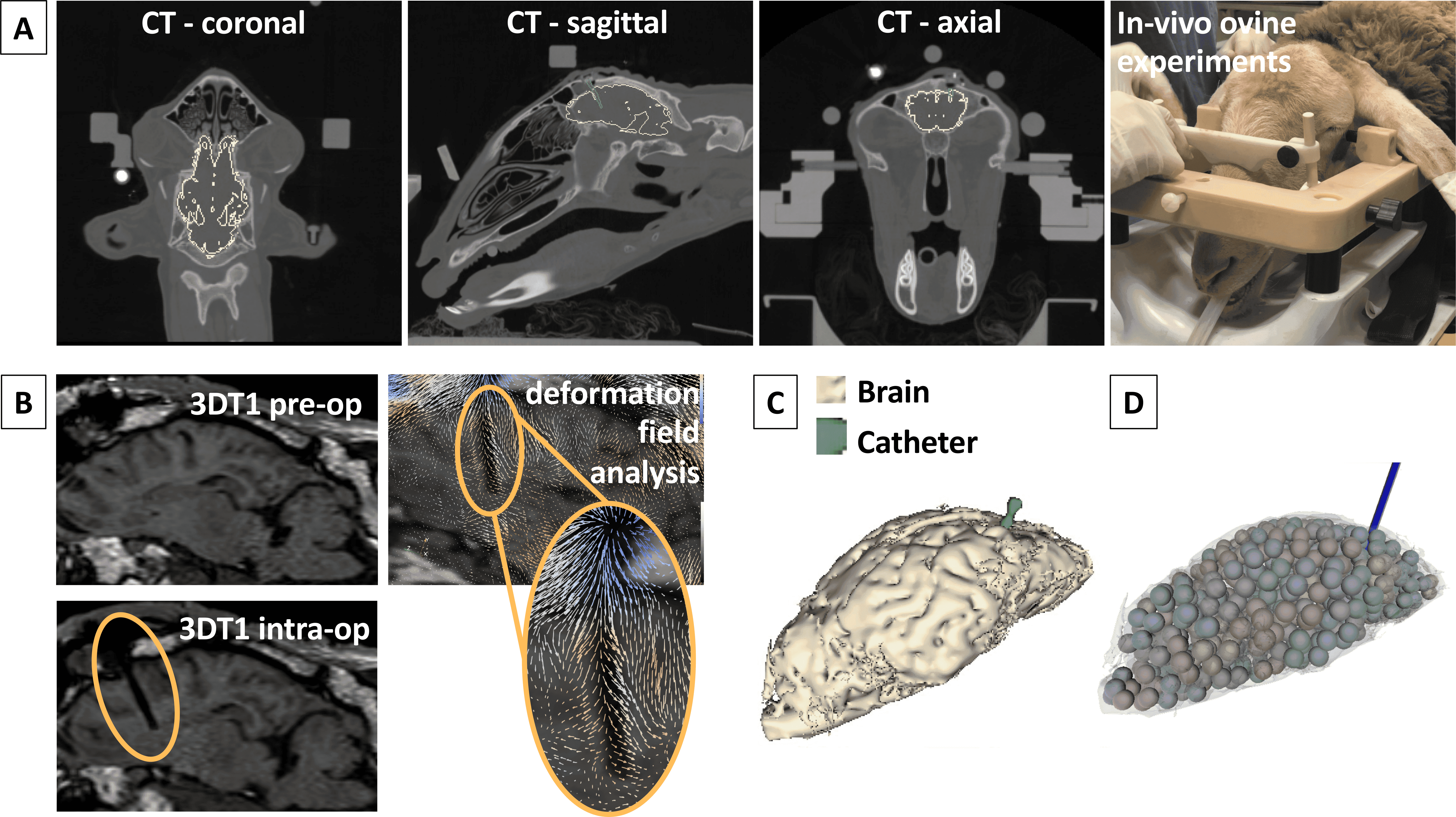} 
    \caption{\textbf{(A)} Coronal, sagittal and axial view of post-operative CT on ovine brain and in-vivo experiment. \textbf{(B)} Ovine brain pre-operative and intra-operative 3DT1 and deformation field analysis. \textbf{(C)} 3D segmented dataset for the ovine brain. \textbf{(D)} Particle model for the ovine brain.}
    \label{fig:figure4}
\end{figure*}
During the experiments carried out in Università degli Studi di Milano Ospedale Veterinario Universitario within the EU’s Horizon EDEN2020 project, five animals underwent \gls{mr} at 1.5 $T$ (Achieva, Philips Healthcare) before and after the procedure (\gls{pbn} insertion), including \gls{ct} (Fig.~\ref{fig:figure4}A) and conventional 3DT1 \gls{ffe} (TR/TE:25/5$ms$) to guide the neurosurgical planning (Fig.~\ref{fig:figure4}B). We segmented the brain on the normalized 3D T1-weighted images, as shown in Fig.~\ref{fig:figure4}C. All animals were treated in accordance with the European Communities Council directive (2010/63/EU), to the laws and regulations on animal welfare enclosed in D.L.G.S. 26/2014. Ethical approval for this study was obtained by the Italian Health Department with authorization n 635/2017.
\subsubsection{Simulation paramenters} 
First, we calibrated the simulation parameters for brain white and deep gray matter with an empirical procedure based on calibration experiments and observations of the resulting visual and numerical deformations. These parameters were then validated by comparing the obtained deformations with those in in-vivo tests on the ovine brain. Catheter-tissue interaction is modeled as a contact problem, handled by the Unity engine's default collision detection and response implementation.
\paragraph{White matter parameters calibration}
We created a scene in Unity containing the previously mentioned parallelepiped-shaped simulated phantom. To describe the entire model as deformable, we constrain all the particles to fall within at least one cluster by imposing \textit{cluster radius} to be at least half of \textit{cluster spacing} (set to [5, 35] $mm$), as proposed in~\cite{tagliabue19}. Consequently, \textit{cluster radius} is restricted to the range [2.5, 35] $mm$ to keep the simulation stable; the upper limit is coincident with the \textit{cluster spacing} one to maintain an overlap between the various clusters. 
Hence, the final deformation is smooth without burdening the computational cost. Conversely, \textit{cluster stiffness} is left free to vary within the entire acceptable range {[0,1]} with a trial and error process to find the most suitable value. Indeed, the model's stiffness does not depend only on the defined stiffness parameter but also on the time step, number of solver iterations, number of clusters, and shape-matching constraints. This involves considering the stiffness parameters in relation to the entirety of the system and highlights how the \gls{pbd} parameters, in general, do not have a direct physical meaning~\cite{camara}.   

To initialize the parameters, we performed 8 insertions of the catheter inside the simulated phantom, $insertion^{k}$ (with $1\leq k \leq 8$). 
Following the experimental work of Leibinger et al.~\cite{imperial4}, we have set the following conditions: the catheter is inserted in a straight trajectory with a constant speed of 0.5~$mm/s$, without offset at the tip. The temporal evolution of all particles' average displacement was evaluated at 31.5~$mm$ depth along the insertion axis. Last, the sampling of the displacement was calculated with a fixed space interval of 3.4~$mm$.

As result, at each frame $f$ of the simulation the following metrics were calculated:
\begin{itemize}
\item The penetration depth ($\Delta_{depth_f}$) of the $k$ insertion:
\begin{equation}
\Delta depth_{f}^k = \left \| \textbf{q}_{init}^k - \textbf{q}_{f}^k \right \| \label{eq:00}
\end{equation}
where $\textbf{q}_{init}^k$ is the catheter's starting $x,y,z$ position, $\textbf{q}_{f}^k$ is the position of the tip of the catheter at frame $f$, and $\|.\|$ is the Euclidean distance.
\end{itemize}
For each interval of $3.4~mm$ we calculated:
\begin{itemize}
\item The average displacement ($\Delta_{disp_f}$) of the N particles of the parallelepiped-shaped phantom at the depth of $31.4~mm$:
\begin{equation}
\Delta disp_{f}^k = \frac{1}{N}\sum \left \| \textbf{p}_{init}^k - \textbf{p}_{f}^k \right \| \label{eq:01}
\end{equation}
where $\textbf{p}_{i,init}^k$ represents the starting position of the N particles, and $\textbf{p}_{i,f}^k$ their position at frame $f$.
\item The displacement ($\Delta_{centerDisp_f}$) of phantom particles \gls{com}:
\begin{equation}
\Delta centerDisp_{f} = \left \| \textbf{c}_{init} - \textbf{c}_{f} \right \| \label{eq:02}
\end{equation}
where $\textbf{c}_{init}^k$ represents the starting position of the particles \gls{com}, and $\textbf{c}_{f}$ their position at frame $f$.
\end{itemize}
We averaged the measured variables across the 8 experiments. Subsequently, starting from the displacement results reported in~\cite{imperial4}, we fine-tuned the \gls{pbd} model parameters to achieve comparable values for the brain white matter.

\paragraph{Deep gray matter parameters calibration}
We tuned the gray matter parameters following the procedure of parameter initialization of the white matter and knowing the differences in the various brain structures' behaviors~\cite{stiffness1,stiffness2,stiffness3,stiffness4,stiffness5,stiffness6}. White matter presents a real shear modulus higher than gray matter. The same applies to the viscosity, represented by the imaginary shear modulus, leading to an isotropic complex shear modulus of white matter around 1.25 $kPa$~\cite{stiffness2}, and a higher stiffness with respect to gray matter. Other brain components were analyzed through a 3D multi-frequency \gls{mre}, and values of the complex shear modulus between 1.058~$kPa$ for the thalamus and 0.649~$kPa$ for the caudate were observed. The stiffness values for the cerebellum and the brainstem are slightly higher, as measured by~\cite{brainshift3}.
According to this possible subdivision of the brain into areas with similar stiffness present in the literature, we tuned the FleX asset parameters of the NVIDIA FleX framework to give a simulation behavior as close as possible to reality.

We created a new Unity scene containing all the previously segmented brain structures to validate the chosen simulation parameters. Fig.~\ref{fig:figure3}B shows the application of the FleX model to gyri and sulci; on the other hand, in Fig.~\ref{fig:figure3}C, it is possible to observe the different brain structures with their flexible particle system.
We selected one desired starting position for the catheter, $\textbf{q}_{init}$, on the brain cortex and a target position, $\textbf{q}_{goal}$, for the left hemisphere and to manually perform an insertion from $\textbf{q}_{init}$ towards $\textbf{q}_{goal}$ using a hand controller. For each $insertion^{k}$ (with $1\leq k \leq 8$), we evaluated the average displacement of each brain structure's centers. 

\begin{figure}
    \centering
    \includegraphics[width=\columnwidth]{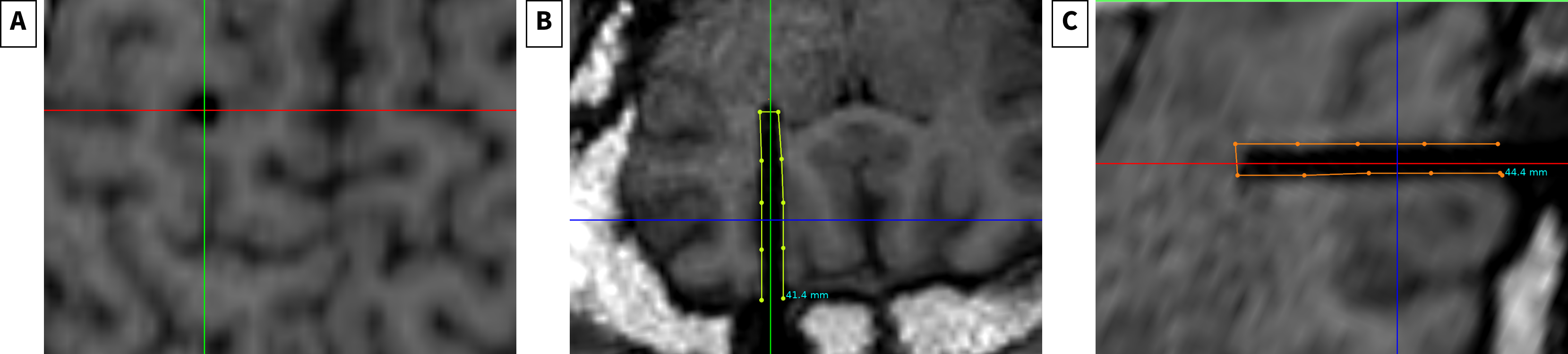}
    \caption{\textbf{(A)} To sample the deformation field, first we set the viewing axes so that the z-axis is roughly parallel to the catheter insertion direction. \textbf{(B)} Then we take our samples from the catheter hole boundary in the x-z plane (green points) \textbf{(C)} and y-z plane (orange point).
    }
    \label{fig:figure5}
\end{figure}

For each frame $f$ of the simulation we calculated:
\begin{itemize}
\item The penetration depth ($\Delta_{depth_f}$) of the $k$ insertion, as defined in Equation~\ref{eq:00};
\item The displacement ($\Delta_{disp_f}$) of the \gls{com} $c$ of each brain structure ($\{BS\}$):
\begin{equation}
\Delta disp_{i,f}^k = \left \| \textbf{c}_{i,init}^k - \textbf{c}_{i,f}^k \right \|\label{eq:002}
\end{equation}
where $\textbf{c}_{i,init}^k$ represents the \gls{com} position of the $i$ $bs$ when the \gls{pbn} tip is at the start position $\textbf{q}_{init}^k$; conversely, $\textbf{c}_{i,f}^k$ represent the \gls{com} position at frame $f$.
\end{itemize}
Subsequently, we computed the mean displacement averaging the displacements of the \glspl{com} in all the frames $f$, all structures $bs_i$, and all the experiments $k$.
\paragraph{Brain deformation model validation}
We computed the displacement of points surrounding the catheter inserted in one ovine brain to validate the calibration parameters. First, using ImFusion Suite~\cite{imfusion}, we rigidly registered the post-operative \gls{mr} (after catheter insertion) to the pre-operative one. After that, we performed a non-rigid registration using the Demons algorithm~\cite{demon} and normalized cross-correlation (NCC) as a similarity metric. To obtain points for evaluation, we aligned the axes to the catheter cylinder on the post-operative image. Subsequently, we extracted the displacement field; on each plane, we picked $z = 1, ... , 5$ equally distanced points on each side $s = 1, ..., 4$ of the hole created from the catheter insertion.
The deformation field is sampled from the perimeter of the cylinder hole created by the insertion, visible on the \gls{mr}. To select points on the perimeter, we set the viewing axes so that the z-axis is in line with the insertion direction, shown in Fig.~\ref{fig:figure5}A. Next, we selected $5$ points on each side of the cylinder hole in $x-z$ plane (green points in Fig.~\ref{fig:figure5}B) and $y-z$ plane (orange points in Fig.~\ref{fig:figure5}C) resulting in $20$ points in total. After that, we simulated the segmented ovine brain into the Unity scene using the particle model (Fig.~\ref{fig:figure4}D). $\forall z, s$ we measured the displacements $\Delta disp_{f}^k$ (Eq.~\ref{eq:01}).
\begin{figure*}
    \centering
    \includegraphics[width=\textwidth]{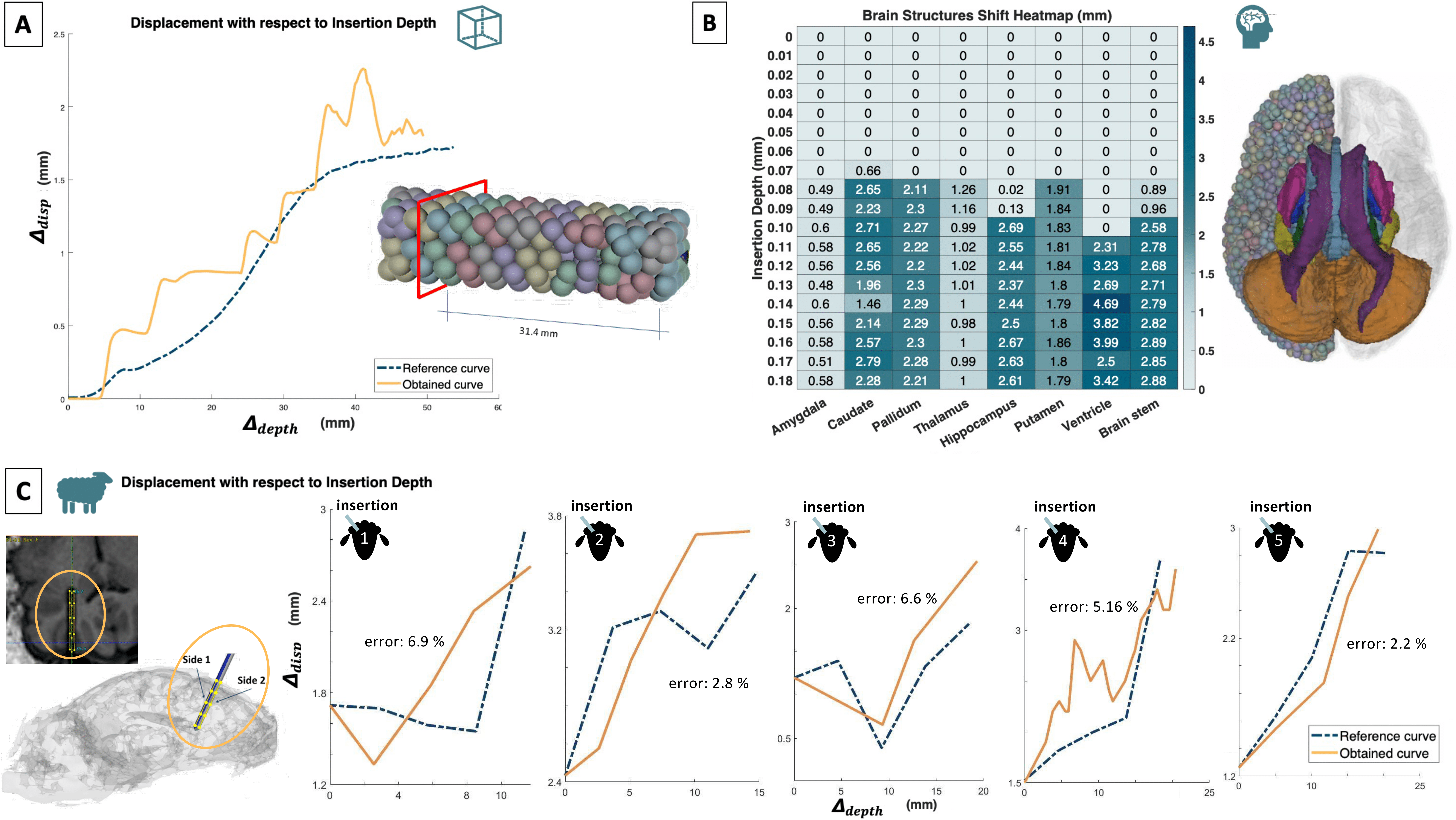}
    \caption{\textbf{(A)} Displacement obtained for white matter on the calibration parallelepiped and comparison with the one obtained in~\cite{imperial4}. \textbf{(B)} Heatmap representing the displacement of the different deep gray matter structures at various insertion depths. \textbf{(C)} Five experiments reporting the deformation of the simulated ovine model (yellow) and the relative in-vivo reference (blue) as a function of the insertion depth.}
    \label{fig:figure6}
\end{figure*}
\subsection{Computational Setup}
The simulation environment was developed and tested in Unity 2019.2 using NVIDIA FleX on a workstation equipped with an Intel Core i7 6800k processor, 32 GB RAM, Titan Xp GPU by NVIDIA Corporation with CUDA 10.1. 
\section{Results}
\subsection{White matter calibration}
This section reports the results related to the calibration of the white matter simulation parameters and their tuning.
The curve shown in Fig.~\ref{fig:figure6}A represents the temporal evolution of the average displacement of all the particles situated at a depth of 31.4~$mm$ within the tissue phantom. We computed the displacements at a fixed interval of 0.34~$mm$ while inserting the catheter. We compared this temporal evolution with the one obtained in the experiment presented in~\cite{imperial4}. We observed that parameter tuning led to a deformation range comparable with the reference one. 
\subsection{Deep gray matter calibration}
\begin{table}[t!]
\centering
\caption{ADIMENSIONAL PARAMETERS SET FOR DIFFERENT BRAIN STRUCTURES AS A RESULT OF THE FINE-TUNING PROCEDURE}
\label{tab:params}
\small\addtolength{\tabcolsep}{-4.5pt}
\begin{tabular}{cccccc}
\textit{\textbf{\begin{tabular}[c]{@{}c@{}}Flex\\ Objects\end{tabular}}} & \textit{\textbf{\begin{tabular}[c]{@{}c@{}}Particle\\ Spacing\end{tabular}}} & \textit{\textbf{\begin{tabular}[c]{@{}c@{}}Cluster Spacing\\ and Radius\end{tabular}}} & \textit{\textbf{\begin{tabular}[c]{@{}c@{}}Cluster\\ Stiffness\end{tabular}}} & \textit{\textbf{\begin{tabular}[c]{@{}c@{}}Link\\ Radius\end{tabular}}} & \textit{\textbf{\begin{tabular}[c]{@{}c@{}}Link\\ Stiffness\end{tabular}}} \\ \hline
Amygdala & 0.005 & 0.005 & 0.0005 & 0.005 & 0.001 \\
Caudate & 0.005 & 0.005 & 0.0005 & 0.005 & 0.001 \\
Pallidum & 0.005 & 0.006 & 0.0005 & 0.005 & 0.001 \\
Hippocampus & 0.005 & 0.005 & 0.0005 & 0.005 & 0.001 \\\hline
Thalamus & 0.005 & 0.005 & 0.001 & 0.005 & 0.001 \\ 
Ventricle & 0.0065 & 0.0065 & 0.001 & 0.005 & 0.001 \\ 
Putamen & 0.0065 & 0.0065 & 0.0005 & 0.005 & 0.001 \\ \hline
Gyri & 0.0066 & 0.0066 & 0.002 & 0.009 & 0.001 \\
Sulci & 0.007 & 0.007 & 0.001 & 0.005 & 0.001 \\ \hline
Brain Stem & 0.007 & 0.007 & 0.0005 & 0.005 & 0.001 \\
Cerebellum & 0.006 & 0.006 & 0.002 & 0.0065 & 0.001
\end{tabular}
\end{table}
Tab.~\ref{tab:params} reports the optimal values for cluster spacing, radius, and stiffness parameters obtained with the fine-tuning procedure for the various brain structures.
Fig.~\ref{fig:figure6}B summarizes the effect of the catheter insertion in the simulated deep gray matter. On the x-axis, we reported the various brain structures, whereas the different insertion depths ({\em i.e.,} the increment on the y-direction starting from the anterior limit of the brain) are represented on the y-axis. Each cell of the grid provides the value of the particular brain structure center's displacement at the relative insertion depth.
The shift of the structures obtained with the calibration reflects the expected values reported in~\cite{imperial4}. This suggests that a proper tuning of the tissue deformation parameters has been performed.
\subsection{Brain deformation model validation}
We confronted the displacement obtained during five different catheter insertions simulated on our ovine model and the ones obtained in the in-vivo experiments for the model validation. Fig.~\ref{fig:figure6}C shows this comparison considering the insertion depth. It can be noticed that the deformation ranges are comparable, as shown by the mean mismatch of 4.73$\pm$2.15\% obtained by computing the mean squared errors of the average displacements (with an average re-planning latency of 0.02 $sec$). Hence, the model can be applied to different types of datasets and represents a good surrogate for the modelization of deformation induced by the catheter insertion in different cerebral areas.
\section{Discussion} \label{sec:discussion}
This paper presents a model able to account for the brain's dynamic behavior during keyhole neurosurgery. 
It is established that the accuracy of \gls{fem} results would benefit from using a higher mesh resolution, but this would come at the expense of degradation in computation time~\cite{zhang}.
Conversely, the mesh-free \gls{pbd} approach has the advantage of avoiding the time-consuming generation of high-quality meshes, which represents the major bottleneck in \gls{fem} simulations (especially in those involving large deformations). Since we target a patient-specific context, this represents a relevant advantage because the mesh would have to be constructed every time for each patient. Furthermore, thanks to its direct manipulation of positions, the \gls{pbd} approach can efficiently handle collision constraints. Probe-tissue interaction can thus be effectively treated as a collision problem, thus allowing to deal with any input probe position without requiring the explicit definition of the contacting surface. The same does not apply to \gls{fem} simulations, where the enforcement of contact constraints would introduce degradation of the performances and stability issues.
The proposed approach relies on the \textit{region-based shape matching} constraint to model large deformations of soft tissues. This implementation's main drawbacks are the dependence of the deformable behavior on time step size and iteration count and the fact that \gls{pbd} parameters do not have a direct physical meaning. The preliminary calibration of the main deformation parameters on a distinct geometry was performed offline to find reasonable initialization values. Before applying the deformation model, simulation parameters were refined with a fine-tuning procedure on the final structures of interest to improve parameter values to describe patient-specific features. Indeed, the \gls{pbd} approach is controlled by a high number of parameters that, if fine-tuned, can provide a more realistic simulation. We have demonstrated that by tuning a subset of parameters on a \gls{mch} phantom can lead to matching real deformations. We obtained a \gls{pbd} model able to replicate the typical behavior of the brain. This is demonstrated by comparing the deformations produced by our model applied to the ovine brain and the corresponding real deformations, obtaining an average error of 4.73\%. This fine-tuning process of all parameters is out of the scope of this work, and it will be explored in future development. We expect that such parameters will be able to model another patient's brain by simply replacing the tuned meshes in the publicly available project with the new specific brain anatomical models. We do not expect that the chosen parameters (given in Table \ref{tab:params}) will be able to model other parts of the brain that were not tuned (\emph{e.g.}, subthalamic nucleus). However, with proper tuning of their values, it will be possible to account for intra-brain variability. Overall, the stability and latency of the proposed are satisfactory. To improve the stability of the simulated environment, it would be optimal to increment the total number of particles (thus increasing the degrees of freedom) or force a particle to lie at the real fiducials' exact location.
\section{Conclusions} \label{sec:conclusion}
Through this work, we have presented a realistic, time-bounded simulator that mocks brain deformations during keyhole surgical procedures, where a catheter/needle is inserted into the brain. The simulator's numerical model has used a novel approach with respect to the literature, and it has proved to be a close match with real brain deformations through validation using recorded deformation data of in-vivo animal trials. The simulator represents a critical component for the development and training of \gls{ai} systems such as autonomous control or intra-operative path planners in the context of \gls{kn}, which will be explored and presented in future works. 

\section*{Acknowledgment}
The authors would like to thank the EDEN2020 project consortium partners for their precious advice during the project activities. We thank A. Castellano and A. Falini for providing the dataset for the simulation.

\ifCLASSOPTIONcaptionsoff
  \newpage
\fi



\bibliographystyle{IEEEtran}
\bibliography{IEEEabrv,main}
%



\end{document}